%% file: acl_latex.tex
\documentclass[11pt]{article}

\usepackage[final]{acl}

\usepackage{times}
\usepackage{latexsym}

\usepackage[T1]{fontenc}

\usepackage[utf8]{inputenc}

\usepackage{microtype}

\usepackage{inconsolata}

\usepackage{graphicx}

\usepackage{hyperref}       
\usepackage{url}            
\usepackage{booktabs}       
\usepackage{amsfonts}       
\usepackage{nicefrac}       
\usepackage{microtype}      
\usepackage{colortbl}

\usepackage{subcaption}
\usepackage{amsmath}
\usepackage{amssymb}
\usepackage{mathtools}
\usepackage{amsthm}

\usepackage{enumitem}
\definecolor{deepblue}{RGB}{0, 0, 139}
\hypersetup{
    colorlinks=true,
    linkcolor=black,
    citecolor=deepblue,
    urlcolor=deepblue
}

\definecolor{lightpink}{rgb}{1.0, 0.95, 0.95}

\usepackage{flushend}

\usepackage{listings}

\title{From Final Artifacts to Trajectories: Retrospective Process Supervision for Evidence-Grounded Long-Form Generation}

\author{
Junjie Huang$^{1}$, Jiarui Qin$^{2}$, Di Yin$^{2}$, Weiwen Liu$^{1}$, \\
\textbf{Yong Yu$^{1}$, Xing Sun$^{2}$, Weinan Zhang$^{1}$} \\
$^{1}$Shanghai Jiao Tong University,\enspace$^{2}$Tencent Youtu Lab \\
\texttt{huangjunjie2019@sjtu.edu.cn},\enspace \texttt{qinjr@icloud.com}
}

\begin{document}
\maketitle
\begin{abstract}
Trajectory data is getting more vital for training large language models for boosting the agentic abilities. 
Unlike the verifiable domains such as coding or mathematics, scaling trajectory data for open-ended tasks is much more difficult because these tasks lack singular ground truth and are costly to annotate or verify.
In this paper, we propose \textsc{RetroGen}, a self-improving framework of retrospective process supervision. 
Our key observation is that although expert trajectories are scarce, high-quality final artifacts such as literature reviews, analyst reports and legal judgments, are abundant in pre-training data and can be \textit{viewed as compressed traces of the evidence-seeking processes that produced them}.  
\textsc{RetroGen} reconstructs candidate latent trajectories from expert artifacts, verifies them against both the artifact and supporting evidence, and trains models on their own successful reconstruction data, \textit{without requiring trajectory data from stronger models}. 
Experiments show that \textsc{RetroGen} improves grounding, faithful synthesis, and long-form evidence-seeking agent tasks.
\end{abstract}

\input{paper/introduction}
\input{paper/preliminary}
\input{paper/method}
\input{paper/experiments}
\input{paper/related_work}
\input{paper/conclusion}

\input{paper/limitation}
\input{paper/ack}
\bibliography{custom}

\newpage

\input{paper/appendix}

\end{document}

%% file: paper/introduction.tex
\section{Introduction}

\begin{figure}[!t]
    \centering
    \includegraphics[width=\linewidth]{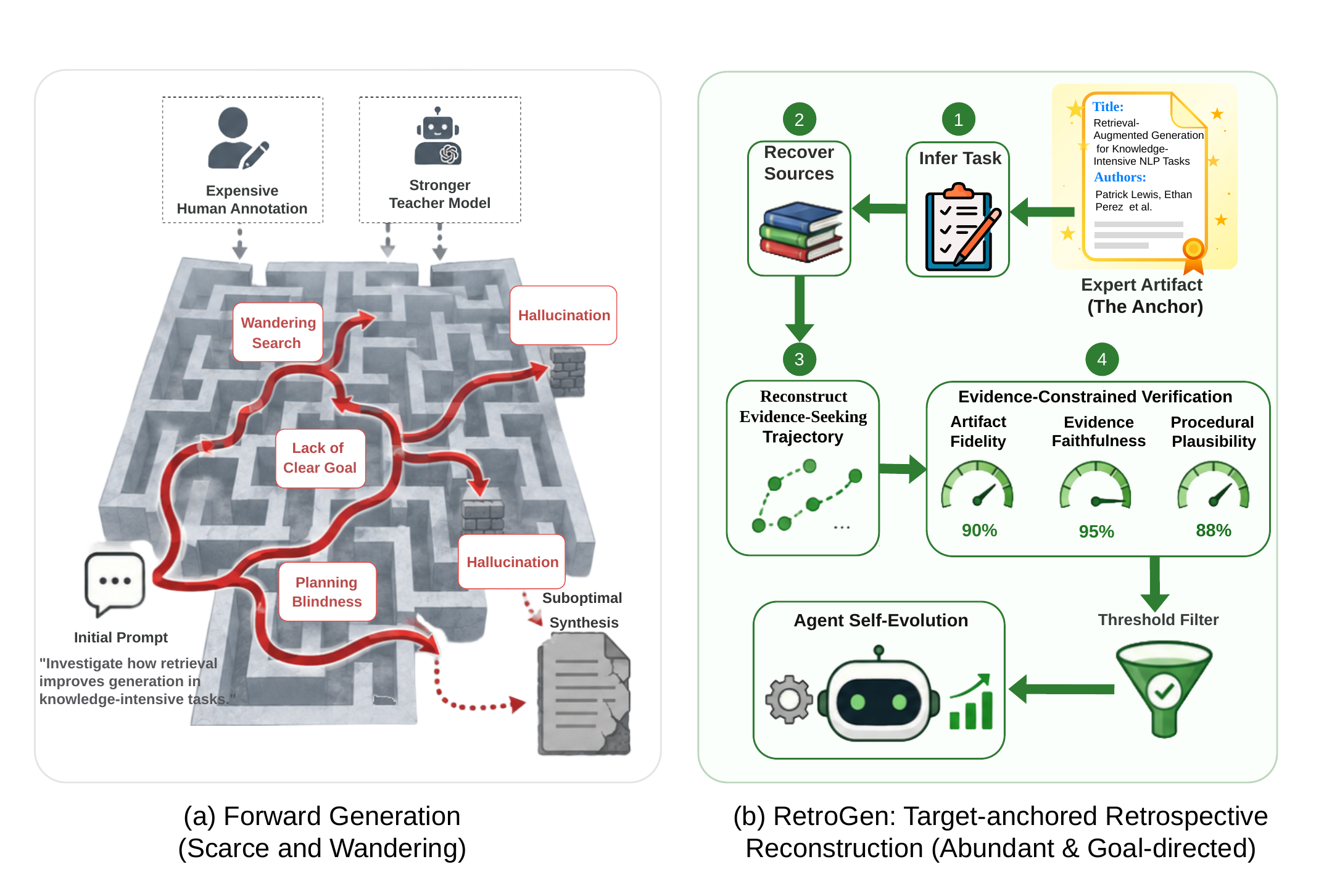}
    \caption{Illustration of our motivation. \textsc{RetroGen} conceptualizes abundant expert artifacts as compressed processes, enabling scalable and grounded process supervision for open-ended agents without costly human or teacher model annotation.}
    \label{fig:intro}
    \vspace{-4mm}
\end{figure}

Large language models (LLMs) are increasingly being extended into autonomous agents capable of planning, tool use, and multi-step interaction~\citep{yao2022react, shinn2023reflexion, wang2024survey, qin2024toolllm}. A key factor driving this progress is trajectory-level supervision: training models not only on final answers, but also on intermediate reasoning, actions, observations, and revisions~\citep{zelikman2022star, lightman2024let, shao2024deepseekmath}. 
In verifiable domains such as mathematics, coding, and closed-ended question answering, generating trajectory data at scale is more feasible because the data can be verified using clear signals.

However, this reliance on objective verification creates a fundamental bottleneck for open-ended, evidence-grounded tasks, such as synthesizing literature reviews~\citep{openscholar,mcdonald2022detect}, composing financial analyst reports grounded in filings~\citep{loukas2021edgar}, or drafting legal judgments~\citep{guha2023legalbench,dai2025laiw}. Unlike closed-ended problems, these tasks lack a singular ground truth that can validate the generated trajectory. Even when a final output appears coherent, it remains difficult to automatically determine whether the selected evidence is sufficient, whether claims are faithfully grounded, and whether the reasoning path follows domain-specific standards of rigor. As a result, scalable process supervision for evidence-grounded long-form generation remains an important open challenge.

Existing approaches typically obtain trajectory supervision by distilling from stronger teacher models or relying on human experts~\citep{liu2023webglm,qin2023webcpm,lightman2024let}. Yet, both sources are inherently unscalable: teacher distillation requires repeated access to costly proprietary models, while human annotation is prohibitively expensive for evidence-intensive workflows. Moreover, relying on forward trajectory generation in open-ended domains often leads to wandering search behaviors, as models lack the holistic foresight required to synthesize complex information.

This motivates a more self-improving, target-driven alternative. As illustrated in Figure~\ref{fig:intro}, we observe that while high-quality open-ended trajectories are exceptionally scarce, high-quality final artifacts produced by domain experts are remarkably abundant. A published literature review or a binding legal judgment is not merely a static output; rather, it is a lossy compression of a latent, multi-step evidence-seeking process. A well-crafted related work section implicitly encodes how an author searched, filtered, compared, and synthesized prior studies into a coherent narrative. Crucially, because these artifacts are rigorously vetted by domain experts (e.g., peer reviewers), they provide strong quality prior for reverse-engineering the processes that produced them.

This perspective transforms expert-curated artifacts into verifiable targets for scalable process supervision. Instead of demanding an external teacher to demonstrate a full forward trajectory, the agent is tasked with reconstructing a plausible evidence-seeking trajectory that arrives at the known expert artifact.

In this paper, we introduce \textsc{RetroGen}, a self-improving framework that operationalizes retrospective process supervision. Given an expert artifact, \textsc{RetroGen} first establishes an evidence environment by retrieving and organizing relevant sources. The agent then reconstructs a trace detailing how the artifact was derived, encompassing evidence selection, claim grounding, and intermediate synthesis. Rather than treating all generated traces as valid supervision, \textsc{RetroGen} employs a rubric-guided verification step that quantitatively scores candidate trajectories across multiple dimensions. By applying a weighted thresholding mechanism, the agent selectively filters reconstructions based on their comprehensive performance in artifact fidelity, evidence faithfulness, and procedural plausibility. This establishes an iterative artifact-to-trajectory-to-agent training loop.

Our contributions are summarized as follows:
\begin{itemize}[leftmargin=*, topsep=1pt, partopsep=0pt, itemsep=0pt, parsep=3pt] 
    \item We formulate expert-curated artifacts as a scalable source of retrospective process supervision, reframing static outputs as verifiable targets for trajectory reconstruction.
    \item We propose \textsc{RetroGen}, an iterative framework that autonomously generates, verifies, and trains on reconstructed trajectories without relying on stronger teacher models.
\end{itemize}

The proposed \textsc{RetroGen} significantly improves evidence-grounded long text generation on both static and agentic benchmarks, while preserving general capabilities. A verification ablation further shows that combining scoring dimensions outperforms any single filter.

%% file: paper/preliminary.tex
\section{Problem Formulation}
\label{sec:problem}

We study \emph{open-ended, evidence-grounded long-form generation} tasks where the final output is an expert-written artifact, but the generative process is unobserved. Given only an artifact $y^{*}$ (e.g., a related-work section, financial analysis report, or legal judgment), our goal is to reconstruct a plausible evidence-seeking trajectory that yields it.  

\paragraph{Latent Expert Process.}
We formulate $y^*$ as the outcome of an unobserved expert workflow:
\begin{equation}
(x^*, \mathcal{E}^*, \tau^*) \mapsto y^*,
\end{equation}
where $x^*$ is the underlying task specification, $\mathcal{E}^{*}$ represents the supporting evidence consulted by the expert, and $\tau^*$ is the latent process transforming the task and evidence into the final artifact. Because these variables are unobserved, we rely on the assumption that $y^{*}$ serves as a lossy but highly informative compression of this latent process.

\paragraph{Retrospective Reconstruction.}
Recognizing that the exact historical process $\tau^{*}$ is unknowable, we instead aim to reconstruct a plausible sequence:
\begin{equation}
\hat{x} = g_x(y^*), \;
\hat{\mathcal{E}} = g_e(y^*), \;
\hat{\tau} = g_\tau(y^*, \hat{x}, \hat{\mathcal{E}}),
\end{equation}
where $\hat{x}$ is the reconstructed task, $\hat{\mathcal{E}}$ is the recovered evidence set, and $\hat{\tau}$ is a reverse-engineered trace describing how an agent could derive an output consistent with $y^*$.
Because the exact historical process $\tau^{*}$ is unobserved, we treat $\hat{\tau}$ as an \emph{executable surrogate trajectory}: a tool-using workflow that is faithful to $y^{*}$ and useful for training, rather than a reconstruction of the expert's unrecorded trial-and-error history.

\paragraph{Desiderata.}
A high-quality reconstructed trace $\hat{\tau}$ should satisfy three properties:

\begin{itemize}[leftmargin=*, topsep=3pt, partopsep=0pt, itemsep=0pt, parsep=3pt] 
    \item Artifact Fidelity: Preserving the essential content and structural organization of $y^{*}$.
    \item Evidence Faithfulness: Grounding intermediate reasoning and final claims strictly within the recovered evidence $\hat{\mathcal{E}}$.
    \item Procedural Plausibility: Exhibiting a logical domain workflow (e.g., search, inspect, compare) rather than hallucinating leaps in logic.
\end{itemize}



%% file: paper/method.tex
\section{Method}
\label{sec:method}

Given an expert artifact $y^{*}$, \textsc{RetroGen} constructs retrospective process-supervision data through a structured pipeline. 

\subsection{Artifact-Anchored Initialization}
Because the forward task is underdetermined, we first infer the starting conditions directly from the expert artifact. Leveraging its inherent reasoning capabilities, the agent infers a plausible task specification $\hat{x}$ that aligns with the final output.

Concurrently, it induces an instance-specific rubric $R=\{(r_{k},c_{k})\}_{k=1}^{K}$ to serve as the evaluation standard for later verification. Rather than relying on generic quality metrics, this rubric extracts fine-grained, checkable criteria ($r_{k}$) mapped to specific target aspects ($c_{k}$) inherent in the artifact. For instance, if the artifact is a legal judgment document, a criterion $r_{k}$ might demand "explicitly invoking the specific tort law precedent", targeting the aspect $c_{k}$ of "statutory grounding". Alternatively, for an academic survey, $r_{k}$ might require "contrasting the computational overhead of two specific baseline models", mapping to $c_{k}$ for "comparative analysis". 

Finally, to establish the evidence environment, we extract concrete cues, such as explicit citations, statutory references, or key entities, from the artifact. These cues seed an artifact-conditioned retrieval over domain-specific corpora, yielding the recovered evidence set $\hat{\mathcal{E}}$. Crucially, this entire initialization phase is fully self-bootstrapped, requiring zero human-authored process traces or external teacher models.

\subsection{Trajectory Reconstruction}
Given $\hat{x}$ and $\hat{\mathcal{E}}$, the agent autonomously formulates a high-level operational plan $\hat{\pi}=(s_{1},s_{2},...,s_{T})$ utilizing a shared abstract action vocabulary $\mathcal{A} = \{\text{Search}, \text{Open}, \text{Extract}, \text{Compare}, \text{Outline}, \text{Draft}\}$. 
This plan is subsequently materialized into candidate tool-augmented traces:
\begin{equation}
\hat{\tau} = (\underbrace{\hat{o}_{0}, \hat{a}_{1}, \hat{o}_{1}, \dots, \hat{n}_{t}}_{\text{denoted as } \hat{\tau}^{\mathrm{pre}}}, \tilde{y}),
\end{equation}
where $\hat{a}_{t} \in \mathcal{A}$ denotes a specific tool invocation, $\hat{o}_{t}$ represents the corresponding environmental observation, and $\hat{n}_{t}$ is an intermediate synthesis note. Collectively, these intermediate elements form the reasoning and evidence-seeking prefix $\hat{\tau}^{\mathrm{pre}}$, and $\tilde{y}$ is the final generated artifact.
Crucially, the observations $\hat{o}_{t}$ are populated via actual tool executions rather than model hallucinations, ensuring that the reconstructed traces are empirically faithful rather than merely narratively plausible.

To prevent the reconstructed dataset from collapsing into a single, homogeneous trace style, we inject controlled diversity across query formulation, plan realization, outline granularity, reflection frequency, and surface formatting. This acts as a form of procedural regularization, ensuring that the model learns robust and generalizable evidence-seeking behaviors instead of merely memorizing a rigid procedural template.

\subsection{Evidence-Constrained Verification}
Each candidate trace $\hat{\tau}$ is evaluated with respect to the desiderata in Section~\ref{sec:problem}: artifact fidelity, evidence faithfulness, and procedural plausibility. We operationalize this through a rubric-guided, multi-dimensional scoring mechanism that yields four complementary signals.

First, a rubric score ($s_{\mathrm{rub}}$) quantifies whether the synthesized artifact $\tilde{y}$ fully satisfies, partially satisfies, or misses each self-induced criterion within $R$.
Second, a holistic quality score ($s_{\mathrm{qual}}$) measures overarching domain rigor, encompassing dimensions such as coherence, factual accuracy, legal correctness, or analytical depth, depending on the specific task.
Third, a grounding score ($s_{\mathrm{grd}}$) ensures that intermediate synthesis notes $\hat{n}_t$ and final claims are strictly supported by the recovered evidence environment $\hat{\mathcal{E}}$.
Fourth, a trace-consistency score ($s_{\mathrm{con}}$) verifies the internal logic of the workflow—checking, for example, whether comparison steps correctly reference previously opened documents, whether outlines are reflected in the final structure, and whether final claims are logically licensed by earlier observations.
The overall trace score is computed as a weighted sum:
\begin{equation}
    \mathrm{Score}(\hat{\tau})=\lambda_r s_{\mathrm{rub}}+\lambda_q s_{\mathrm{qual}}+\lambda_g s_{\mathrm{grd}}+\lambda_c s_{\mathrm{con}},\end{equation} where $\lambda_{\{r,q,g,c\}}$ denote the respective mixing weights. Traces exhibiting failed retrievals, malformed tool executions, or overall scores falling below a domain-specific threshold are automatically discarded.
    Crucially, this weighted thresholding mechanism does not attempt to certify that a trace is historically identical to the original human workflow. Instead, it acts as a scalable proxy for process quality, ensuring that the retained trajectories are faithful to the expert anchor, grounded in evidence, and procedurally robust, all without requiring human oversight.

\subsection{Retrospective Process Supervision}

After the evidence-constrained verification, we retain the successfully filtered candidate traces. Crucially, rather than forcing the model to predict the exact original expert artifact $y^*$ at the end of the sequence, we pair the retained trace prefix $\hat{\tau}^{\mathrm{pre}}_i$ with its corresponding synthesized draft $\tilde{y}_i$. Concretely, each training sample encompasses the inferred task specification, the recovered evidence context, the tool-interaction trace, the intermediate notes, and the verified synthetic artifact:\begin{equation}(\hat{x}_i, \hat{\mathcal{E}}_i, \hat{\tau}^{\mathrm{pre}}_i, \tilde{y}_i).\end{equation}Using the synthesized output $\tilde{y}_i$ instead of the original $y_i^*$ ensures strict causal consistency between the intermediate reasoning steps and the final generation. Because the trace has already passed our rigorous multi-dimensional verification—which guarantees high fidelity to the original artifact—$\tilde{y}_i$ maintains high quality for the agent to learn from.
We serialize each sample into a single autoregressive sequence and optimize the model using standard language modeling:
\begin{equation}
z_i = \mathrm{Serialize}(\hat{x}_i, \hat{\mathcal{E}}_i, \hat{\tau}^{\mathrm{pre}}_i, \tilde{y}_i),
\end{equation}
\begin{equation}
\mathcal{L}(\theta)=\sum_i \sum_t\log p_\theta(z_{i,t} \mid z_{i,<t}).\end{equation}
Compared with final-output-only supervision, this self-improving objective teaches the model not only what high-quality artifact to produce, but also how to autonomously decompose an open-ended task, gather and organize evidence, and synthesize it into a rigorously grounded long-form output.

%% file: paper/experiments.tex
\section{Experiments}

\input{tables/main_result}
\input{tables/verification_ablation}

\subsection{Experimental Setup}
\label{sec:setup}

\paragraph{Domains and Artifacts.}
We evaluate \textsc{RetroGen} on three evidence-grounded long-form generation domains: 
(i) \textbf{scientific writing}, where the task is to draft a related-work section conditioned on a paper abstract and the expert artifact $y^*$ is the author-written related-work section from the corresponding paper;
(ii) \textbf{financial analysis}, where the agent produces an analyst-style report grounded in U.S. SEC 10-K filings from the EDGAR corpus~\citep{loukas2021edgar} and $y^*$ is a human-written financial narrative; and
(iii) \textbf{legal judgment drafting}, where the agent drafts a first-instance judgment grounded in statutes and prior cases, with $y^*$ given by the official adjudicated judgment.
These domains stress-test RetroGen under distinct evidence regimes: citation-heavy scientific synthesis, document-grounded financial reasoning, and statute/case-grounded legal argumentation.

\paragraph{Tools and Execution Environment.}
All trajectories are generated using domain-specific instantiations of the abstract action space $\mathcal{A}$ defined in Section~\ref{sec:method}. To ensure that supervision reflects executable evidence-seeking behavior, observations are populated through actual tool calls rather than model-imagined snippets. We cap each trajectory at $30$ tool steps and maintain an $8{,}000$-character sliding context window.

\paragraph{Backbone Models.}
We evaluate \textsc{RetroGen} on four open-source LLMs: Qwen3-8B~\citep{yang2025qwen3}, Qwen2.5-7B~\citep{Yang2024Qwen25TR}, Mistral-7B-v0.3~\citep{Jiang2023Mistral7}, and Olmo-3-1025-7B~\citep{olmo2025olmo}. 

\paragraph{Training Corpus.}
For each backbone, \textsc{RetroGen} reconstructs candidate trajectories from expert artifacts and filters them using the multi-dimensional verification score in Section~\ref{sec:method} with uniform weights $\lambda=0.25$. Thresholds are calibrated per domain from the score distribution (0.52 / 0.48 / 0.63 for scientific / financial / legal; Appendix~\ref{app:verification}), retaining approximately $25$K verified trajectories. We construct a $\sim$50M-token SFT corpus with a fixed token-budgeted mixture: $80\%$ verified agentic trajectories, split $40$:$20$:$20$ across scientific, financial, and legal domains, and $20\%$ general instruction-following, math, coding, and multi-turn dialogue data. This mixture preserves general capabilities while emphasizing evidence-seeking behavior.

\paragraph{Baselines and Ablations.}
We compare against controlled alternatives under the same token budget, sequence length, and optimization configuration:
\begin{enumerate}[label=(\arabic*), leftmargin=*, topsep=3pt, partopsep=0pt, itemsep=0pt, parsep=3pt]
    \item \textbf{Initial}: the checkpoint before training.
    \item \textbf{ForwardGen}: trajectories generated by the same backbone from the task specification $\hat{x}$ without access to the expert artifact, isolating the value of target-anchored retrospective reconstruction.
    \item \textbf{Artifact-Only}: the same reconstructed dataset with all intermediate tool interactions removed, isolating trajectory-level process supervision from input-output supervision.
    \item \textbf{Public-Baseline}: the $80\%$ agentic slice is replaced with WebGLM-QA~\citep{liu2023webglm} and WebCPM-WK~\citep{qin2023webcpm}, two public citation-grounded long-form generation corpora that provide retrieve-then-write supervision distilled from stronger systems.
\end{enumerate}

\paragraph{Evaluation.}
We evaluate along three axes. First, \emph{evidence-grounded long-form generation} is measured by ALCE~\citep{gao2023enabling}, ScholarQA~\citep{openscholar}, and QASPER~\citep{mcdonald2022detect}. Second, \emph{domain-specific grounded reasoning} is assessed with legal reasoning benchmarks including LAiW~\citep{dai2025laiw} and LegalBench~\cite{guha2023legalbench}. Third, \emph{general capability retention} is measured by GSM8K~\citep{cobbe2021gsm8k}, MMLU-Pro~\citep{wang2024mmlu}, IFEval~\citep{zhou2023instruction}, and HumanEval~\citep{chen2021codex}. All evaluations use vLLM-backed zero/few-shot inference~\citep{kwon2023efficient}, with details deferred to Appendix.

\subsection{Main Results}
\begin{figure*}[!t]
    \centering
    \includegraphics[width=0.8\linewidth]{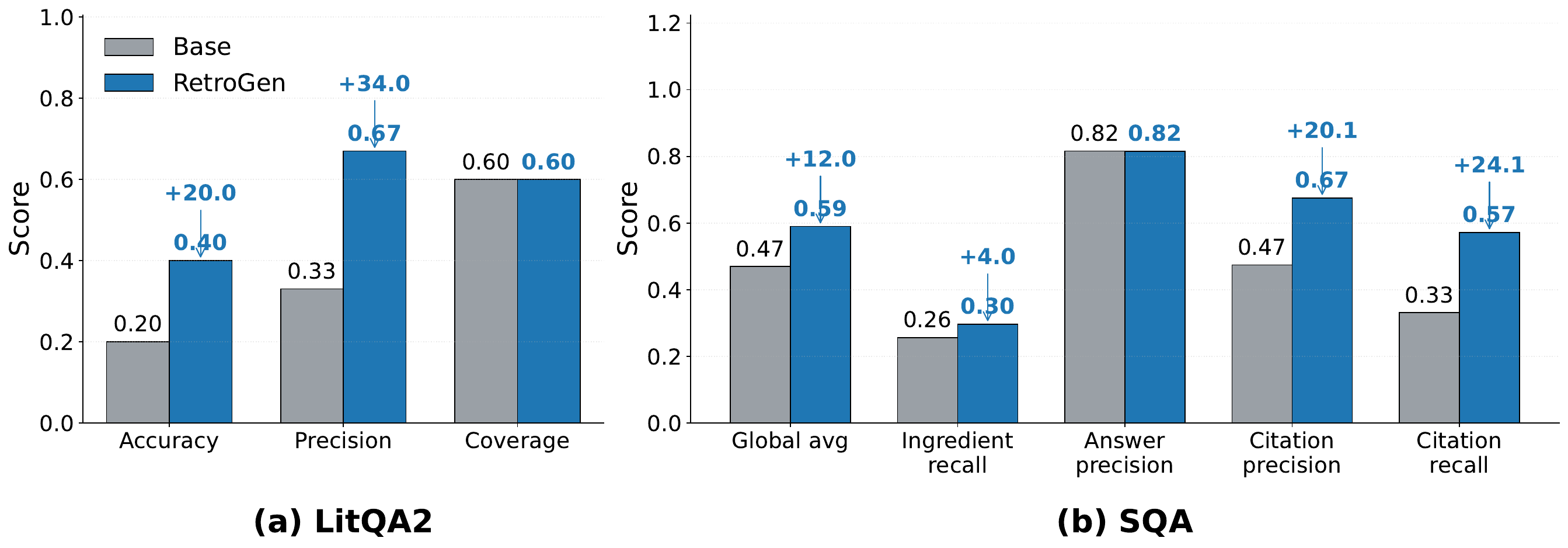}
    \caption{AstaBench literature-understanding performance on LitQA2 and SQA. }
    \label{fig:exp_asta}
\end{figure*}

\begin{figure*}[!t]
    \centering
    \includegraphics[width=0.8\linewidth]{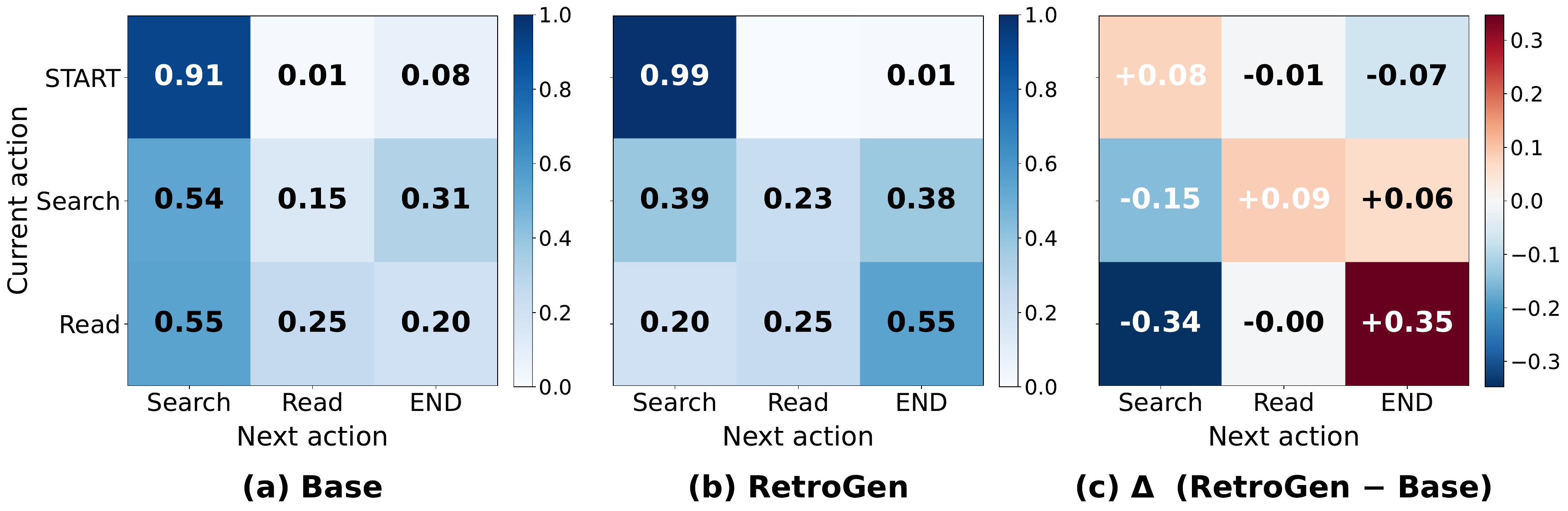}
    \caption{AstaBench action-transition matrices, where each cell denotes the probability of transitioning from the current action to the next action.}
    \label{fig:exp_trans}
    \vspace{-3mm}
\end{figure*}

Table~\ref{tab:main_results} reports the performance of \textsc{RetroGen} and all baselines across the four backbone models. We summarize three main observations.

\begin{itemize}[leftmargin=*]
    \item \textbf{Retrospective anchoring improves evidence-grounded generation.}
Across ALCE, ScholarQA, and QASPER, \textsc{RetroGen} consistently outperforms the Initial and ForwardGen models across backbones. Compared with ForwardGen, which generates trajectories from the task specification $\hat{x}$ without access to the expert artifact, \textsc{RetroGen} improves the average evidence-grounded generation score by +4.8 points, with gains of +4.3 on ALCE, +6.7 on ScholarQA, and +3.5 on QASPER. This suggests that anchoring trajectory reconstruction on expert artifacts helps reduce unguided exploration and produces more reliable evidence-seeking behavior.

\item \textbf{Trajectory-level supervision provides gains beyond final-artifact imitation.}
The comparison with Artifact-Only isolates the value of intermediate tool-interaction traces. On domain-specific reasoning benchmarks, \textsc{RetroGen} improves over Artifact-Only by +2.7 points on average, with gains of +3.4 on LAiW and +2.0 on LegalBench. This indicates that final artifacts alone do not fully expose the procedural structure needed for grounded reasoning; reconstructing how evidence is searched, selected, and organized provides additional supervision for complex synthesis tasks.

    \item \textbf{Verified agentic trajectories are more effective than collapsed retrieve-then-write supervision.}
    \textsc{RetroGen} also outperforms the Public-Baseline, where the same token budget is allocated to public web-grounded long-form QA data in a retrieve-then-write format. The gains are most pronounced on tasks requiring multi-step evidence aggregation, suggesting that supervision over explicit tool-use trajectories teaches the model more than observing only a retrieved context and final answer.
\end{itemize}

\subsection{In-Depth Analysis}

\subsubsection{Verification-Dimension Ablation}
\label{sec:verifier_ablation}

The main results use a uniformly weighted combination of the four verification signals. To test whether this combination is necessary, we retrain Qwen3-8B under the same token budget while filtering with a single score, or while replacing the selected traces with the next-highest-scoring ones (\emph{next-$N$ replace}).
Table~\ref{tab:verifier_ablation} shows that the weighted verifier achieves the best average. Single-dimension filters remain above the initial checkpoint, so moderately scored traces can still help; however, next-$N$ replace underperforms full \textsc{RetroGen}, indicating that usefulness is graded and that the threshold trades quality against data volume.

\subsubsection{Dynamic Agentic Evaluation on AstaBench}
\label{sec:astabench}
\begin{figure}[!t]
    \centering
    \includegraphics[width=\linewidth]{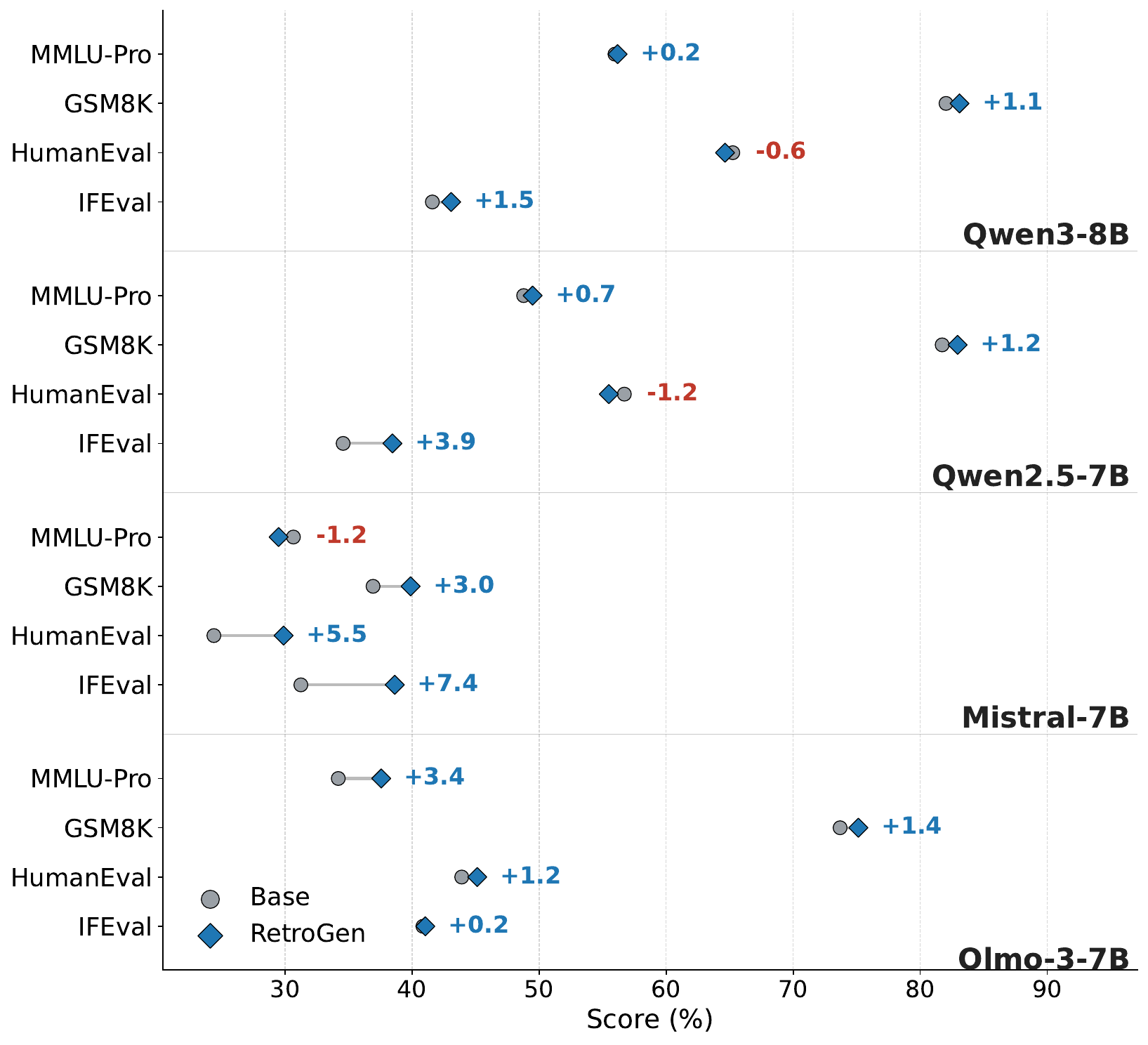}
    \caption{General capability retention across four backbone models. Dumbbell markers compare each initial checkpoint with its \textsc{RetroGen}-trained counterpart.}
    \label{fig:general_cap}
    \vspace{-3mm}
\end{figure}

Standard evidence-grounded generation benchmarks evaluate the quality of final
answers, but they only partially reveal whether a model can behave as a
reliable research agent. To further examine the process-level behavior learned
by \textsc{RetroGen}, we evaluate on AstaBench~\citep{bragg2025astabench}, a
dynamic agentic benchmark for scientific research tasks. Unlike static QA
benchmarks, AstaBench requires the model to iteratively search for papers,
inspect retrieved content, and decide when sufficient evidence has been
collected. This makes it well suited for testing whether retrospective process
supervision improves the agent's evidence-seeking policy rather than only its
final-output style.

\paragraph{Benchmark and protocol.}
We focus on the literature-understanding tasks LitQA2-Validation and SQA (details in Appendix~\ref{app:eval}).
The initial model and its \textsc{RetroGen}-trained
counterpart use the same backbone, tool inventory, sampling configuration, and
maximum interaction budget (\texttt{max\_rounds}$=15$).

\paragraph{Performance on literature understanding.}
Figure~\ref{fig:exp_asta} shows that \textsc{RetroGen} improves most
AstaBench literature-understanding metrics. On LitQA2, \textsc{RetroGen}
doubles accuracy from $0.20$ to $0.40$ and improves precision from $0.33$ to
$0.67$, while maintaining the same coverage of $0.60$. This indicates that the
model becomes more accurate and selective without simply abstaining more often.
On SQA, \textsc{RetroGen} improves the global average from $0.47$ to $0.59$.
The largest gains come from citation-related metrics: citation precision
increases from $0.47$ to $0.67$, and citation recall increases from $0.33$ to
$0.57$. Answer precision remains unchanged at $0.82$, suggesting that the main
benefit is not merely more fluent answer generation, but better grounding and
attribution of claims to supporting evidence.

\paragraph{Action-transition analysis.}
To understand where these gains come from, we further analyze the action
transitions recorded by the AstaBench sandbox. We collapse raw tool calls into
three high-level action types: \textsc{Search}, \textsc{Read}, and \textsc{END}.
Figure~\ref{fig:exp_trans} compares the empirical transition matrices of the
initial model and \textsc{RetroGen}. The initial model frequently repeats search
actions after searching, with a search-to-search probability of $0.54$,
suggesting redundant exploration. In contrast, \textsc{RetroGen} reduces this
probability to $0.39$ and increases the search-to-read transition from $0.15$
to $0.23$, indicating that it is more likely to inspect retrieved evidence
rather than continue issuing new searches. After reading, \textsc{RetroGen}
also transitions to \textsc{END} much more often, increasing the read-to-end
probability from $0.20$ to $0.55$.

Together, these patterns suggest that \textsc{RetroGen} learns a more
goal-directed research policy: it searches to identify candidate evidence,
reads to verify and extract support, and terminates once sufficient evidence has
been gathered. This provides process-level evidence for our main claim that
retrospective supervision improves the structure of evidence-seeking behavior,
not only the quality of the final generated response.

\subsubsection{General Capability Retention}
\label{sec:general_capability}
\begin{figure}[!t]
    \centering
    \includegraphics[width=\linewidth]{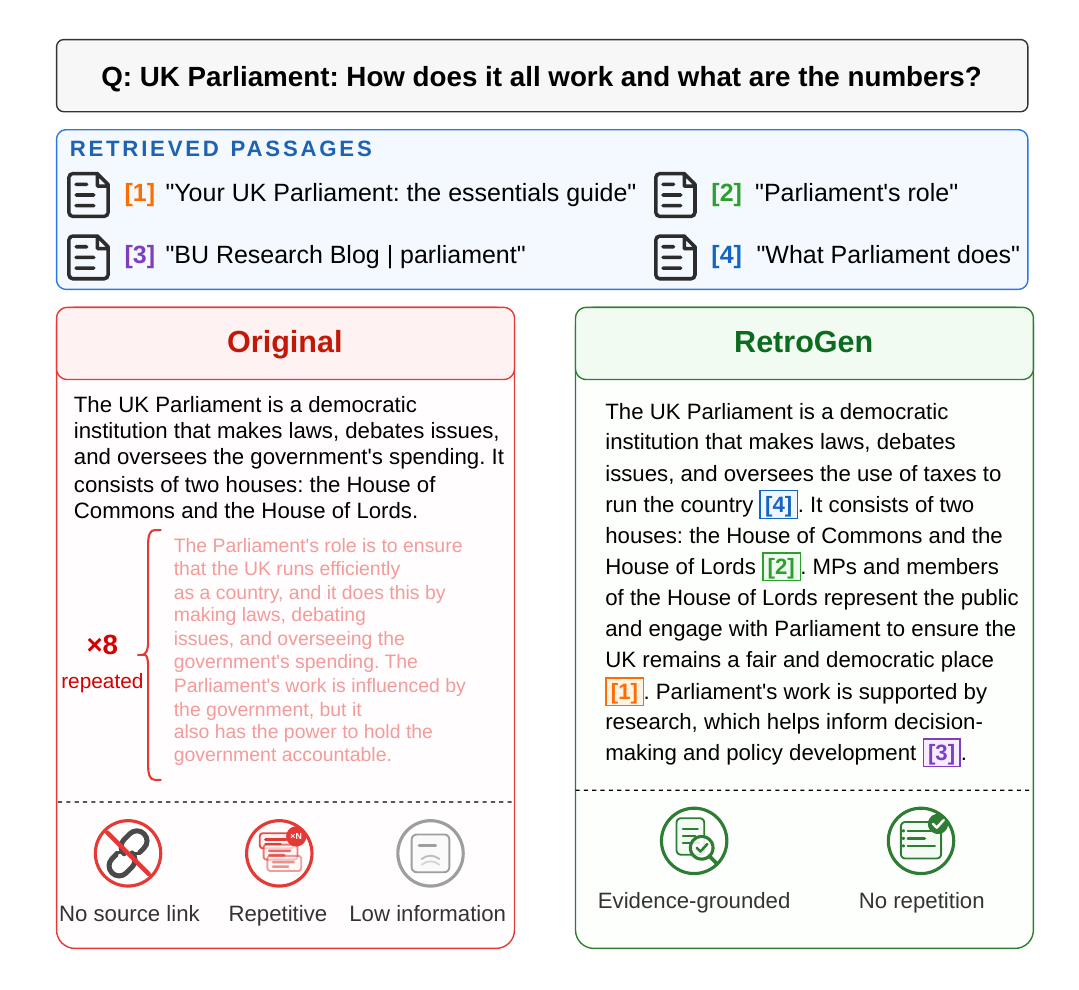}
    \caption{Case study under the same question. \textsc{RetroGen}
produces a more evidence-grounded, less repetitive, and higher-density answer.}
    \label{fig:case_study}
    \vspace{-3mm}
\end{figure}

A potential concern with training on long agentic trajectories is that the model
may become overly specialized to evidence-seeking workflows and lose general
instruction-following or reasoning abilities. Figure~\ref{fig:general_cap} compares each
initial checkpoint with its \textsc{RetroGen}-trained counterpart on GSM8K, MMLU-Pro, IFEval, and HumanEval.
Across backbones, \textsc{RetroGen} shows no systematic degradation: most points are comparable to or slightly above the initial checkpoint. Mixed with a modest amount of general SFT data, verified agentic trajectories therefore improve grounded long-form behavior without inducing catastrophic forgetting.

\subsubsection{Case Study}
\label{sec:case_study}

Figure~\ref{fig:case_study} compares the initial model and \textsc{RetroGen} on the
same question, \emph{``UK Parliament: How does it all work and what are the
numbers?''}. The initial model generates a
plausible overview, but does not attach source links and quickly degenerates
into repeated statements, resulting in low information density.

In contrast, \textsc{RetroGen} produces a more compact and grounded synthesis.
It links claims to the retrieved passages, covers multiple aspects of the
question---including Parliament's role, its two houses, public representation,
and research-supported policymaking---and avoids repetitive phrasing. This
case study illustrates that retrospective process supervision improves not only
evidence use, but also the organization and density of the final answer.

%% file: tables/main_result.tex
\begin{table*}[!t]
\centering
\setlength{\tabcolsep}{4pt}
\renewcommand{\arraystretch}{1.08}
\resizebox{\textwidth}{!}{
\begin{tabular}{
lccccc
@{\hspace{0.6em}} c @{\hspace{0.6em}}
lccccc
}
\toprule

\multicolumn{6}{>{\columncolor{lightpink}}c}{\textbf{Qwen3-8B}}
&
&
\multicolumn{6}{>{\columncolor{lightpink}}c}{\textbf{Mistral-7B-v0.3}} \\
\cmidrule(lr){1-6} \cmidrule(lr){8-13}

\textbf{Method}
& \textbf{ALCE} & \textbf{ScholarQA} & \textbf{QASPER} & \textbf{LAiW} & \textbf{LegalBench}
&
&
\textbf{Method}
& \textbf{ALCE} & \textbf{ScholarQA} & \textbf{QASPER} & \textbf{LAiW} & \textbf{LegalBench} \\

\midrule
Initial & 83.33 & 66.36 & 53.25 & 34.14 & 74.93
&
&
Initial & 68.58 & 27.54 & 54.45 & 20.82 & 53.3 \\

ForwardGen & 85.18 & 65.42 & 51.84 & 33.52 & 74.95
&
&
ForwardGen & 62.41 & 48.65 & 51.34 & 32.15 & 60.27 \\

Artifact-Only & 87.86 & 67.18 & 52.52 & 34.69 & 74.93
&
&
Artifact-Only & 63.64 & 55.83 & 50.21 & 34.82 & 62.22 \\

WebGLM & 84.52 & 65.84 & 53.36 & 36.61 & \textbf{76.12}
&
&
WebGLM & 65.11 & 53.27 & 50.81 & 39.41 & 61.25 \\

WebCPM & 46.73 & 60.27 & 49.89 & 23.23 & 74.92
&
&
WebCPM & 57.90 & 50.41 & 54.40 & 38.68 & 63.10 \\

\textbf{RetroGen (Ours)} & \textbf{88.03} & \textbf{67.71} & \textbf{56.15} & \textbf{37.65} & 75.37
&
&
\textbf{RetroGen (Ours)} & \textbf{68.93} & \textbf{61.12} & \textbf{56.27} & \textbf{41.29} & \textbf{65.98} \\

\midrule

\multicolumn{6}{>{\columncolor{lightpink}}c}{\textbf{Qwen2.5-7B}}
&
&
\multicolumn{6}{>{\columncolor{lightpink}}c}{\textbf{Olmo-3-1025-7B}} \\
\cmidrule(lr){1-6} \cmidrule(lr){8-13}

\textbf{Method}
& \textbf{ALCE} & \textbf{ScholarQA} & \textbf{QASPER} & \textbf{LAiW} & \textbf{LegalBench}
&
&
\textbf{Method}
& \textbf{ALCE} & \textbf{ScholarQA} & \textbf{QASPER} & \textbf{LAiW} & \textbf{LegalBench} \\

\midrule

Initial & 85.37 & 56.42 & 47.03 & 34.71 & 72.16
&
&
Initial & 90.70 & 53.86 & 55.17 & 39.24 & 63.57 \\

ForwardGen & 86.27 & 58.36 & 49.62 & 34.18 & 71.32
&
&
ForwardGen & 89.34 & 55.42 & 53.62 & 37.42 & 64.84 \\

Artifact-Only & 87.48 & 62.74 & 50.04 & 33.39 & 72.41
&
&
Artifact-Only & 91.42& 58.17 & 54.36 & 38.27 & 66.18 \\

WebGLM & 84.26 & 59.85 & 50.15 & 35.24 & 71.27 
&
&
WebGLM & 91.27 & 56.78 & 52.97 & 36.61 & 64.25\\

WebCPM & 67.75 & 54.27 & 48.85 & 31.95 & 72.80
&
&
WebCPM & 74.17 & 53.41 & 51.84 & 34.71 & 64.25 \\

\textbf{RetroGen (Ours)} & \textbf{90.50} & \textbf{64.18} & \textbf{51.64} & \textbf{36.00} & \textbf{75.01}
&
&
\textbf{RetroGen (Ours)} & \textbf{92.83} & \textbf{61.78} & \textbf{56.42} & \textbf{39.65} & \textbf{67.30} \\

\bottomrule
\end{tabular}
}
\caption{Main results across four open-source backbone models. 
\textsc{RetroGen} achieves the best average performance across evidence-grounded generation and domain-specific reasoning benchmarks.}
\label{tab:main_results}
\end{table*}

%% file: tables/verification_ablation.tex
\begin{table*}[!t]
\centering
\small
\setlength{\tabcolsep}{6pt}
\renewcommand{\arraystretch}{1.05}
\begin{tabular}{lcccccc}
\toprule
\textbf{Filter} & \textbf{ALCE} & \textbf{ScholarQA} & \textbf{QASPER} & \textbf{LAiW} & \textbf{LegalBench} & \textbf{Avg} \\
\midrule
Initial & 83.33 & 66.36 & 53.25 & 34.14 & 74.93 & 62.40 \\
only $s_{\mathrm{qual}}$ & 84.88 & 66.62 & 53.48 & 34.72 & 75.46 & 63.03 \\
only $s_{\mathrm{con}}$ & 85.42 & 66.18 & 53.86 & 35.64 & 75.22 & 63.26 \\
next-$N$ replace & 86.42 & 66.74 & 54.18 & 35.86 & 75.12 & 63.66 \\
only $s_{\mathrm{grd}}$ & 86.18 & 66.52 & 55.42 & 36.48 & 75.44 & 64.01 \\
only $s_{\mathrm{rub}}$ & 88.18 & 67.82 & 55.28 & 36.05 & 74.96 & 64.46 \\
\textbf{RetroGen} ($\lambda{=}0.25$) & 88.03 & 67.71 & \textbf{56.15} & \textbf{37.65} & 75.37 & \textbf{64.98} \\
\bottomrule
\end{tabular}
\caption{Verification-dimension ablation on Qwen3-8B under a matched token budget. Uniformly weighted scoring yields the best average; single-signal filters and next-$N$ replacement remain above the initial checkpoint but are less stable.}
\label{tab:verifier_ablation}
\end{table*}

%% file: paper/related_work.tex
\section{Related Work}

\paragraph{Evidence-grounded long-form generation.}
Retrieval-augmented generation improves the factuality and verifiability of language models by grounding generation in external evidence~\citep{lewis2020retrieval}. Recent work extends this paradigm to web-grounded and citation-grounded long-form generation, where models must synthesize evidence from multiple sources and provide faithful attributions~\citep{liu2023webglm,qin2023webcpm,gao2023enabling}. Scientific and legal domains further stress these abilities, as reliable generation often requires long-document reading, evidence comparison, and well-supported argumentation~\citep{openscholar,mcdonald2022detect,guha2023legalbench,dai2025laiw}. However, most existing pipelines supervise models through final answers, retrieved contexts, or distilled responses, collapsing the intermediate evidence-seeking process into a single input-output mapping. \textsc{RetroGen} instead treats expert-written artifacts as compressed traces of latent evidence-seeking processes and reconstructs explicit tool-use trajectories from them.

\paragraph{Tool-use agents and process supervision.}
A growing line of work studies language models as agents that interleave reasoning with external actions such as search, retrieval, API calls, and environment interaction~\citep{wang2024survey,yao2022react,qin2024toolllm}. While tool use extends models beyond parametric knowledge, training reliable agents remains challenging because high-quality process trajectories are scarce. Process supervision shows that intermediate reasoning steps can improve reliability over outcome-only supervision~\citep{lightman2024let}, and reasoning bootstrapping suggests that models can benefit from learning from their own intermediate rationales~\citep{zelikman2022star,shao2024deepseekmath}. Unlike these settings, evidence-seeking agents require not only plausible reasoning chains but also faithful observations produced by executable tools. \textsc{RetroGen} therefore reconstructs candidate trajectories retrospectively and filters them with evidence-constrained verification.

\paragraph{Self-improvement without stronger teachers.}
Self-improving agents have been explored through verbal feedback, reflection, and iterative refinement~\citep{shinn2023reflexion}. However, many data-generation pipelines still rely on stronger teacher models, human demonstrations, or preference labels to produce supervision at scale. Web-grounded systems such as WebGLM and WebCPM show the value of retrieval-enhanced supervision for long-form QA~\citep{liu2023webglm,qin2023webcpm}, but typically organize supervision around retrieve-then-write outputs rather than reusable multi-step agentic processes. \textsc{RetroGen} differs by using expert-curated artifacts as target anchors: instead of asking a stronger model or human annotator to generate trajectories forward, it infers plausible evidence-seeking trajectories backward from the artifact, verifies them automatically, and trains the backbone on the resulting retrospective process supervision.

%% file: paper/conclusion.tex
\section{Conclusion}

We presented \textsc{RetroGen}, a framework for retrospective process supervision that trains evidence-seeking agents from expert-curated final artifacts. Rather than collecting costly human process annotations or relying on stronger teacher models, \textsc{RetroGen} reconstructs candidate tool-use trajectories from high-quality artifacts and filters them through evidence-constrained verification for artifact fidelity, evidence faithfulness, and procedural plausibility.

Across scientific writing, financial analysis, and legal judgment drafting, \textsc{RetroGen} improves evidence-grounded long-form generation and domain-specific reasoning over various baselines. Ablations show that the four verification signals are complementary, and in-depth analyses further show that the method induces more structured agentic behavior on dynamic research tasks while preserving general capabilities. These findings suggest that expert artifacts can serve not only as final targets, but also as scalable sources of process supervision for self-improving open-ended agents.

%% file: paper/limitation.tex
\section*{Limitations}

\textsc{RetroGen} relies on the availability of high-quality expert artifacts. While such artifacts are often more abundant than process annotations, their usefulness depends on whether they contain enough recoverable evidence and structural signals to support retrospective reconstruction. In domains where final outputs are highly underspecified, stylistically diverse, or weakly grounded in explicit evidence, the reconstructed trajectories may be less reliable.

Our experiments focus on evidence-grounded long-form generation with retrieval-oriented tools. Although the framework is designed to be general, further work is needed to evaluate retrospective process supervision in domains requiring richer interactive environments, longer-horizon planning, or non-textual tools such as code execution, database operations, and multimodal perception.

%% file: paper/ack.tex
\section*{Acknowledgments}

The work is supported by National Natural Science
Foundation of China (62502310,62322603).

%% file: paper/appendix.tex
\newpage
\appendix

\section{Implementation Details}
\label{app:impl}

\subsection{Expert Artifact Corpora}
\label{app:corpora}

For each of the three domains in Section~\ref{sec:setup}, the expert
artifact $y^\ast$ is drawn from a publicly available source and used as
the anchor for retrospective reconstruction.

\begin{itemize}[leftmargin=*]
  \item \textbf{Scientific writing.}
    We use author-written related-work sections as expert artifacts
    $y^\ast$. Each instance consists of a parsed related-work block
    together with the corresponding paper abstract and cited reference
    list. The abstract is used as side context for inducing the task
    specification $\hat{x}$, while the original related-work text is
    never exposed to the agent during executable trajectory generation.

  \item \textbf{Financial analysis.}
    We use Item~7 (Management's Discussion and Analysis, MD\&A)
    sections of U.S. SEC 10-K filings from the EDGAR corpus
    \citep{loukas2021edgar}. Each MD\&A section is treated as the
    expert artifact $y^\ast$ and is indexed by company and fiscal year.
    The remaining filing content, such as business description, risk
    factors, and quantitative disclosures, is available to the agent
    only through actual filing-reader tool calls.

  \item \textbf{Legal judgment drafting.}
    We use Chinese first-instance civil and administrative judgments
    collected from publicly accessible court-record portals. Each
    judgment is normalized into plain text, with personal identifiers
    pseudonymized when necessary. Statutes and prior cases are not
    bundled with $y^\ast$ and must be retrieved at trajectory time
    through the legal search and browse tools.
\end{itemize}

Across all domains, the target artifact is used only as a retrospective
anchor for task induction, evidence recovery, and verification. It is not
available as an observation during trajectory execution.

\subsection{Procedural Diversity}
\label{app:diversity}

To prevent the reconstructed corpus from collapsing into a single
template, we randomize five axes when sampling each trajectory:
(i) user-query phrasing;
(ii) system prompt style, including variations in tone, persona, and
tool schema;
(iii) high-level plan flow pattern;
(iv) thought templates for each action type, such as \textsc{Search},
\textsc{Browse}, \textsc{Synthesize}, \textsc{Assemble},
\textsc{Reflection}, \textsc{Transition}, and \textsc{ErrorRecovery};
and (v) the surface formatting of tool-call blocks, such as JSON-style
tool calls and ReAct-style \texttt{Action/Action Input} formatting. All
random seeds are fixed within a single backbone run for reproducibility.

\subsection{SFT Mixture and Training}
\label{app:sft}

The same SFT recipe is applied to all fine-tuned variants. The
\textbf{Base} baseline denotes the original checkpoint without SFT.

\paragraph{Token-budgeted mixture.}
We treat the training corpus as a budget of $5\times10^{7}$ tokens. For
\textsc{RetroGen}, this budget is split as:
$40\%$ \textsc{Trajectory-Scientific},
$20\%$ \textsc{Trajectory-Financial},
$20\%$ \textsc{Trajectory-Legal}, and
$20\%$ \textsc{General}. The \textsc{General} pool aggregates
instruction-following, multi-turn dialogue, long-form writing, math,
and code samples. Samples longer than $32{,}768$ tokens are dropped.
The same token budget and mixture proportions are used for the
corresponding baselines, with only the $80\%$ agentic slice changed
according to the baseline definition.

\paragraph{Baseline construction.}
For \textbf{ForwardGen}, trajectories are generated from the inferred
task specification $\hat{x}$ without access to the expert artifact
$y^\ast$ as a target anchor. For \textbf{Artifact-Only}, we remove
intermediate tool calls, observations, and synthesis notes while
preserving the inferred task and final artifact. For
\textbf{Public-Baseline}, we replace the $80\%$ agentic trajectory
slice with WebGLM-QA and WebCPM-WK examples while keeping the same
$20\%$ general-data mixture. All fine-tuned variants share the same
sequence length cap and optimization configuration.

\paragraph{Optimization.}
Training is implemented with full-parameter supervised fine-tuning
using DeepSpeed ZeRO-1 and sequence packing. We use AdamW with peak
learning rate $1\times10^{-5}$, cosine scheduling, warmup ratio $5\%$,
weight decay $0.01$, gradient clipping $1.0$, gradient checkpointing,
left truncation, dropout off, and a target effective batch size of
approximately $512$K tokens per optimizer step. Each backbone is
trained for one epoch with a fixed random seed. The same optimization
configuration is used for every fine-tuned baseline.

\subsection{Evaluation Details}
\label{app:eval}

All evaluations are run with a vLLM backend.
We use \texttt{max\_length}=32{,}768 for evidence-grounded and legal
tasks and $4{,}096$ otherwise. Unless otherwise specified, the random
seed is fixed to $1234$. 

For dynamic agentic evaluation in Section~\ref{sec:astabench}, we run
AstaBench's literature-understanding suite, including
LitQA2-Validation and SQA. LitQA2 measures accuracy, precision, and
coverage on biomedical evidence-seeking questions. SQA reports
ingredient recall, answer precision, citation precision, citation
recall, and their global average. The base model and its
\textsc{RetroGen}-trained counterpart use the same backbone, tool
inventory, sampling configuration, and maximum interaction budget
(\texttt{max\_rounds}=15). Judge-based metrics are rescored using the
same local judge model for all systems.

\subsection{AstaBench Action Abstraction}
\label{app:astabench_actions}

For the transition analysis in Figure~\ref{fig:exp_trans}, we collapse
raw AstaBench tool calls into three high-level actions:
\textsc{Search}, \textsc{Read}, and \textsc{END}. \textsc{Search}
includes paper search, snippet search, citation expansion, and other
retrieval-oriented calls. \textsc{Read} includes opening paper
metadata, reading paper content, and inspecting retrieved passages.
\textsc{END} denotes the model's decision to stop tool use and produce
the final answer. We compute transition probabilities by counting each
consecutive action pair and normalizing by the total outgoing
transitions from the current action.

\subsection{Action Schema and Reusable Templates}
\label{app:schema}

For all three domains, an assistant turn may contain one or more tool
calls followed by an optional free-form note. A final-answer turn emits
no tool call. The default schema is:
\begin{lstlisting}[basicstyle=\ttfamily\scriptsize,breaklines=true]
<tool_call>{"name":"<tool>","arguments":{...}}</tool_call>
\end{lstlisting}
Tool results are returned in the next user turn:
\begin{lstlisting}[basicstyle=\ttfamily\scriptsize,breaklines=true]
<tool_response>...</tool_response>
\end{lstlisting}
We additionally sample a ReAct-style variant at fixed probability per
trajectory:
\begin{lstlisting}[basicstyle=\ttfamily\scriptsize,breaklines=true]
Action: <tool>
Action Input: {...}
\end{lstlisting}
This improves robustness to different tool-call surface forms. Thought
templates that bracket each action are drawn from a large pool of
paraphrases to avoid wording collapse during SFT.

\section{Verification Protocol}
\label{app:verification}

Verification is performed by the same backbone that generates the
candidate trace, under a rubric-guided scoring protocol; we do not
introduce a stronger external judge. Each retained trace receives four
scores in $[0,1]$---rubric satisfaction $s_{\mathrm{rub}}$, holistic
quality $s_{\mathrm{qual}}$, evidence grounding $s_{\mathrm{grd}}$, and
trace consistency $s_{\mathrm{con}}$---which are averaged with uniform
weights $\lambda=0.25$.

\paragraph{Domain-specific thresholds.}
Table~\ref{tab:score_thresholds} reports the score distribution used to
calibrate filters. Rather than applying a global cutoff, we remove the
low-score tail in each domain while retaining enough data for the
50M-token SFT mixture. The resulting thresholds are $0.52$ / $0.48$ /
$0.63$ for scientific / financial / legal traces, yielding approximately
$25$K verified trajectories.

\input{tables/score_thresholds}

\paragraph{Failure modes.}
Rejected traces typically exhibit missed evidence themes, weak
grounding, shallow comparison, or incomplete finals. For example, one
rejected scientific trajectory (overall score $0.30$, rubric score
$0.00$) is asked to write a related-work section on adaptive
optimization under partial observability, but produces fluent prose
about uncertainty and decision-making while omitting the required
themes. This is a common rejection pattern: a plausible-sounding
workflow that recovers the wrong evidence lineage. Accepted traces, by
contrast, search and browse in a noisy retrieval space, retain relevant
sources, discard weakly related results, and synthesize an artifact that
covers the self-induced rubric.

\section{Prompts and Anti-Leakage}
\label{app:prompts}

We include simplified prompt templates used for rubric extraction and
task induction in the scientific-writing domain. Financial and legal
domains follow the same structure with domain-specific criteria
(coverage of filings or statutes, analytical depth, and citation of
controlling authority).

\vspace{2.5pt}
\noindent\textbf{Rubric extraction.}\\[0.6ex]
The scientific-writing prompt template is:
\vspace{0.55ex}
\begin{lstlisting}
Analyze the following Related Work section and extract a structured
evaluation rubric.

## Related Work Section
{related_work}

## Instructions
Identify specific quality criteria that a reconstructed trajectory
should satisfy. Each item should be categorized as one of:
coverage | structure | synthesis | citation

Output JSON:
{
  "rubric_items": [
    {
      "id": "R1",
      "category": "coverage|structure|synthesis|citation",
      "description": "Specific requirement description"
    }
  ]
}
\end{lstlisting}
An example extracted item is:
\begin{lstlisting}
{
  "id": "R1",
  "category": "synthesis",
  "description": "Compare greedy frameworks for adaptive
optimization with Stochastic Depletion problems."
}
\end{lstlisting}

\paragraph{Task induction with anti-leakage.}
The inferred query must specify a realistic professional intent without
leaking the evidence the agent is expected to recover:
\begin{lstlisting}[basicstyle=\ttfamily\scriptsize,breaklines=true]
Based on the following Related Work section, generate a task
description for an AI agent that needs to write a similar Related
Work section using only web search.

## Related Work Section
{related_work}
## Rubric
{rubric}

## Instructions
1. Infer what paper this Related Work belongs to.
2. Specify the domain and key themes.
3. Do NOT reveal specific paper titles/authors, target citations,
   exact conclusions, or evidence that the agent must recover
   via search.
4. Provide enough high-level guidance for search.
\end{lstlisting}
A representative inferred task is: \emph{write a related-work section on
trust-aware recommender systems and matrix factorization, covering
fundamental factorization algorithms, social-network structure, and
trust propagation.} The prompt names the research intent and topical
scope, but still requires the agent to search, inspect, select, and
synthesize evidence. During trajectory execution the original artifact
$y^{*}$ is used only offline as a retrospective anchor.

%% file: tables/score_thresholds.tex
\begin{table}[t]
\centering
\small
\setlength{\tabcolsep}{5pt}
\resizebox{\linewidth}{!}{%
\begin{tabular}{lccccc}
\toprule
\textbf{Domain} & \textbf{Mean} & \textbf{Median} & \textbf{P25} & \textbf{P75} & \textbf{Thr.} \\
\midrule
Scientific & 0.571 & 0.594 & 0.412 & 0.743 & 0.52 \\
Financial & 0.508 & 0.517 & 0.273 & 0.738 & 0.48 \\
Legal & 0.624 & 0.668 & 0.480 & 0.799 & 0.63 \\
\bottomrule
\end{tabular}%
}
\caption{Domain-wise distribution of the overall verification score and the corresponding filtering threshold. Thresholds cut the low-score tail while retaining enough traces for the 50M-token SFT mixture, yielding approximately 25K verified trajectories.}
\label{tab:score_thresholds}
\end{table}